\documentclass[conference]{IEEEtran}
\IEEEoverridecommandlockouts
\usepackage{cite}
\usepackage{amsmath,amssymb,amsfonts}
\usepackage{algorithmic}
\usepackage{graphicx}
\usepackage{textcomp}
\usepackage{xcolor}
\usepackage{url}
\usepackage{array}
\newcolumntype{C}[1]{>{\centering\arraybackslash}m{#1}}
\newcolumntype{P}[1]{>{\centering\arraybackslash}p{#1}}
\usepackage{hyperref}
\hypersetup{
    colorlinks=true,
    linkcolor=black,
    filecolor=black,      
    urlcolor=black,
    citecolor=black,
}
\def\BibTeX{{\rm B\kern-.05em{\sc i\kern-.025em b}\kern-.08em
    T\kern-.1667em\lower.7ex\hbox{E}\kern-.125emX}}

\usepackage{fancyhdr}
\begin{document}

\title{\vspace{8pt}Unimodal and Multimodal \\ Static Facial Expression Recognition for \\Virtual Reality  Users with EmoHeVRDB \\
\thanks{}
}

\author{\IEEEauthorblockN{1\textsuperscript{st} Thorben Ortmann}
\IEEEauthorblockA{
\textit{University of the West of Scotland,} \\
\textit{Hamburg University of Applied Sciences}\\
Hamburg, Germany \\
thorben.ortmann@haw-hamburg.de}
\and
\IEEEauthorblockN{2\textsuperscript{nd} Qi Wang}
\IEEEauthorblockA{
\textit{University of the West of Scotland}\\
Paisley, United Kingdom \\
qi.wang@uws.ac.uk}
\and
\IEEEauthorblockN{3\textsuperscript{rd} Larissa Putzar}
\IEEEauthorblockA{\textit{Hamburg University of Applied Sciences}\\
Hamburg, Germany \\
larissa.putzar@haw-hamburg.de}
}

\maketitle
\thispagestyle{fancy}

\begin{abstract}
In this study, we explored the potential of utilizing Facial Expression Activations (FEAs) captured via the Meta Quest Pro Virtual Reality (VR) headset for Facial Expression Recognition (FER) in VR settings.
Leveraging the EmojiHeroVR Database (EmoHeVRDB), we compared several unimodal approaches and achieved up to 73.02\% accuracy for the static FER task with seven emotion categories.
Furthermore, we integrated FEA and image data in multimodal approaches, observing significant improvements in recognition accuracy.
An intermediate fusion approach achieved the highest accuracy of 80.42\%,
significantly surpassing the baseline evaluation result of 69.84\% reported for EmoHeVRDB's image data. 
Our study is the first to utilize EmoHeVRDB’s unique FEA data for unimodal and multimodal static FER, establishing new benchmarks for FER in VR settings.
Our findings highlight the potential of fusing complementary modalities to enhance FER accuracy in VR settings, where conventional image-based methods are severely limited by the occlusion caused by Head-Mounted Displays (HMDs).
\end{abstract}


\section{Introduction}
Virtual Reality (VR) technology has evolved to offer deeply immersive and interactive experiences that can elicit strong emotional responses \cite{somarathnaVirtualRealityEmotion2022}.
Its applications span a wide range of fields, including therapy, training, education and entertainment \cite{halbigSystematicReviewPhysiological2021}.
Automatic emotion recognition enables the evaluation and dynamic adaption of VR content to match users' emotional states.
Emotion-aware VR applications can make therapeutic sessions more responsive, enhance learning through real-time adaptation to frustration levels, or adjust entertainment content based on user engagement \cite{marin-moralesEmotionRecognitionImmersive2020}.
Facial expressions are a natural and expressive way to convey emotions \cite{ekman2006darwin}.
However, they are rarely used for automatic emotion recognition in VR settings because Head-Mounted Displays (HMDs) occlude the wearer's upper face half, making conventional image-based Facial Expression Recognition (FER) approaches ineffective \cite{ortmannFacialEmotionRecognition2023}.
Recently, we introduced the EmojiHeroVR Database (EmoHeVRDB), the first database focused on HMD-occluded faces \cite{ortmannEmojiherovr2024}.
Besides images, it contains Facial Expression Activations (FEAs) captured via the Meta Quest Pro VR headset.
Our work is the first to examine the potential of EmoHeVRDB's FEA data by leveraging it for the categorical static FER task.
We will first use the FEA data exclusively and then combine it with the image data for a multimodal FER approach.

\section{Related Work}
In VR settings, researchers primarily rely on physiological measures such as Electrocardiography (ECG), Electrodermal Activity (EDA), and Electroencephalography (EEG) to assess users' emotions automatically \cite{halbigSystematicReviewPhysiological2021}.
Only 5 \cite{philippSocialityFacialExpressions2012, pallaviciniVirtualRealityAlways2013, bianFrameworkPhysiologicalIndicators2016, chiricoEffectivenessImmersiveVideos2017, granatoEmpiricalStudyPlayers2020a} of the 42 studies included in Marín-Morales et al.'s review on emotion recognition in VR assessed facial expressions, all using Electromyography (EMG) to capture facial muscle activity \cite{marin-moralesEmotionRecognitionImmersive2020}.
Similarly, EMG was the most common measure in our review on facial emotion recognition in VR, followed by recordings from near-infrared cameras embedded in HMDs \cite{ortmannFacialEmotionRecognition2023}. 
Additionally, a few studies \cite{yongEmotionRecognitionGamers2019, georgescuRecognizingFacialExpressions2019a, houshmandFacialExpressionRecognition2020c, georgescuTeacherStudentTraining2021, georgescuTeacherStudentTrainingTriplet2021a, gotsmanValenceArousalEstimation2021} leveraged existing FER image databases to experiment with artificially HMD-occluded data, reporting accuracy decreases from approximately 59\% to 49\% and 85\% to 82\% on the AffectNet \cite{mollahosseiniAffectNetDatabaseFacial2019a} and FER+ \cite{barsoumFERPLUSTrainingDeep2016} datasets, respectively.
Building upon these findings, we collected data from 37 participants who posed facial expressions while playing our VR game EmojiHeroVR \cite{ortmannEmojiherovr2024}.
Subsequently, we constructed EmoHeVRDB based on 1,778 manually labeled central-view images.
\begin{figure}[hb]
  \centering
    \begin{minipage}{.38\linewidth}
    \centering
    \includegraphics[width=\linewidth]{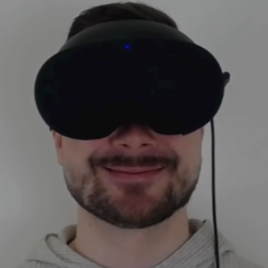}
  \end{minipage}
  \hspace{0.1\linewidth}
  \begin{minipage}{.38\linewidth}
    \centering
    \includegraphics[width=\linewidth]{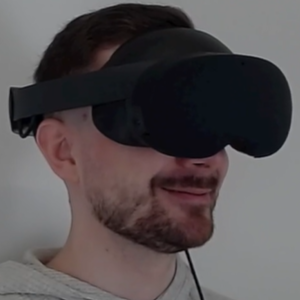}
  \end{minipage}
  \caption{Exemplary central-view and side-view images of the happiness class from EmoHeVRDB.}
  \label{fig:leander}
\end{figure}
As a result, EmoHeVRDB comprises 3,556 image sequences recorded from two angles and 1,727 FEA sequences captured via the Meta Quest Pro VR headset.
The data are categorized into the seven emotion classes: anger, disgust, fear, happiness, neutral, sadness, and surprise.
Using the manually labeled central-view images of EmoHeVRDB and their 45° side-view counterparts, we achieved 69.84\% recognition accuracy on EmoHeVRDB's test set \cite{ortmannEmojiherovr2024}.
However, the potential of EmoHeVRDB's FEA data remained unexplored.

\section{Unimodal Facial Expression Recognition}
\begin{figure}[hb]
  \centering
  \includegraphics[width=.8\linewidth]{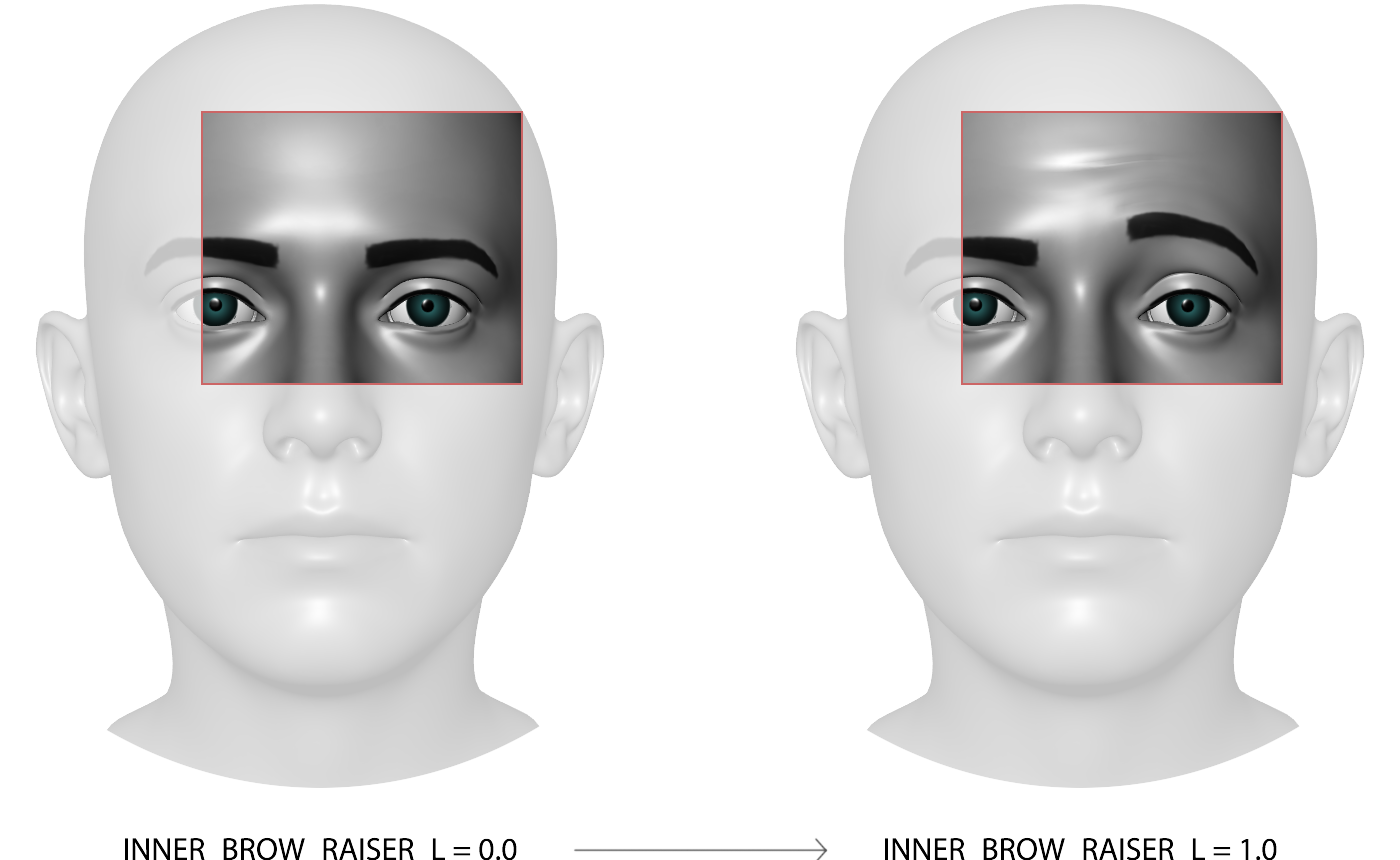}
  \caption{Face blend shape for the \textit{left inner brow raiser} from the Face Tracking API's documentation \cite{meta_face_tracking}.}
  \label{fig:blend}
\end{figure}

The FEAs included in EmoHeVRDB result from calls to the Face Tracking API \cite{meta_face_tracking} of the Meta XR Core SDK, version 59.0 \cite{meta_core_sdk}.
The Face Tracking API relies on the Meta Quest Pro's five inward-facing infrared cameras to detect facial movements.
Each API call returns an array of 63 floating point numbers ranging from 0 to 1.
Each number indicates the activation strength of one of 63 facial expressions, such as \textit{jaw drop} or \textit{left inner brow raiser}, defined based on the Facial Actions Coding System (FACS) \cite{1370848662448368390}.
As visualized in Fig.\ref{fig:blend}, these measurements are typically used to adjust blend shapes to accurately map a user’s facial expressions onto a 3D model.

\subsection{Database Preparation}
We constructed EmoHeVRDB based on 1,778 manually labeled central-view images, each representing a different emotion reenactment process \cite{ortmannEmojiherovr2024}.
In parallel with the image data, FEAs were captured.
Each of the 1,727 FEA sequences in EmoHeVRDB corresponds to the same emotion reenactment process as one of the labeled central-view images.
The number of FEA sequences is lower than the number of manually labeled central-view images because of a technical malfunction of the Meta Quest Pro during our user study to collect the data for EmoHeVRDB.
\begin{figure}[ht]
  \centering
  \includegraphics[width=\linewidth]{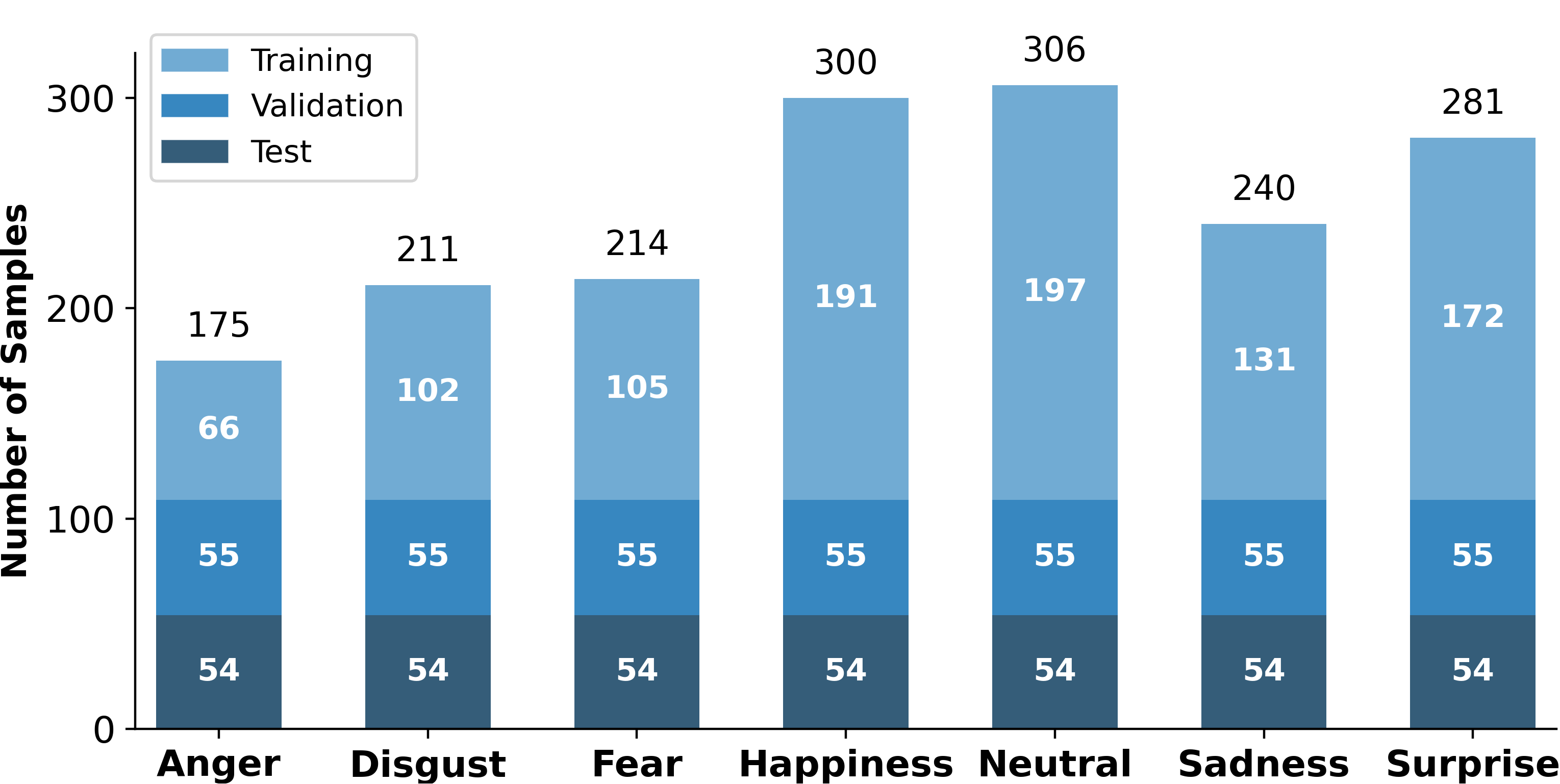}
  \caption{Sample distribution for EmoHeVRDBs FEA data.}
  \label{fig:label-dis}
\end{figure}
For our experiments on static FER, we selected all FEA samples captured simultaneously with a corresponding manually labeled central-view image.
Consequently, we used a subset of EmoHeVRDB comprising 1,727 FEAs divided into seven emotion categories.
We adhered to EmoHeVRDB's predefined participant-independent split, resulting in an unbalanced training set of 964 samples and balanced validation and test sets of 385 and 378 samples, respectively.
The neutral category had the highest class frequency, with 197 samples, while anger had the lowest, with only 66 samples in the training set.
The complete sample distribution is depicted in Fig.\ref{fig:label-dis}.

\subsection{Training}
To leverage the selected FEAs for FER, we needed to train a model that classifies 63-dimensional vectors of floating point numbers ranging from 0 to 1 into exactly one of seven known emotion categories.
Various approaches are available for this task.
Using the scikit-learn machine learning library, version 1.4.2, we experimented with logistic regression, support vector machines (SVMs), random forests, and gradient boosting.
We performed an extensive automated grid search for each model type to identify the optimal hyperparameters, fitting up to 1,620 model candidates per search.
Each model candidate was trained on the training set and evaluated against the validation set to identify the optimal hyperparameters.
An essential hyperparameter for most models was whether to account for the training set's class imbalance in their learning function.
Almost all models benefitted from weighting samples with the inverse of their class frequency.
However, most hyperparameters were model-type-specific, such as the choice of the solver algorithm for logistic regression or the number of trees in the random forest.
Furthermore, we manually optimized the hyperparameters for a multi-layer perceptron (MLP) we built using Tensorflow, version 2.15.
We primarily varied the learning rate, the number of Dense layers and their units, and the number of Dropout layers and their rates. For all experiments, we used a batch size of 32, the Adam optimizer, and the SparseCategoricalCrossentropy loss function.
After hyperparameter optimization, we selected the model that achieved the highest validation accuracy per model type and evaluated it against the test set to determine its performance.
The results are listed in Table \ref{table:performance_metrics}.

\subsection{Results}
The models' performances in terms of accuracy and F-score are very similar, with less than a 3\% difference between the best model, a logistic regression classifier, and the worst model, a gradient boosting classifier.
\begin{table}[ht]
\centering
\normalsize
\caption{Performance Metrics in \% - FEA-based Models}
\begin{tabular}{|l|c|c|c|c|c|}
\hline
\textbf{Model} & \multicolumn{2}{c|}{\textbf{Accuracy}} & \textbf{F1} & \textbf{Min} & \textbf{Min} \\
 \textbf{Type} & \textbf{Val} & \textbf{Test} &  & \textbf{F1} & \textbf{Recall}\\ \hline
LR & \textbf{82.34}  & \textbf{73.02} & \textbf{72.56} & 56.47 & 44.44 \\ 
SVM & 80.52 & 71.43 & 70.90 & 56.82 & 46.30 \\
RF & 82.08 & 71.16 & 69.78 & 46.58 & 31.48 \\
GB & 80.52 & 70.90 & 70.23 & 49.38 & 37.04 \\
MLP & 81.56 & 71.69 & 71.10 & \textbf{57.14} & \textbf{48.15} \\ \hline \hline 
EffNet \cite{ortmannEmojiherovr2024} & 70.39 & 69.84 & 70.06 & 49.11 & 50.93 \\ \hline
\end{tabular}
\label{table:performance_metrics}
\begin{flushleft}
\footnotesize
LR=Logistic Regression, SVM=Support Vector Machine, RF=Random Forest, GB=Gradient Boosting, MLP=Multilayer Perceptron, EffNet=EfficientNet-B0.
\end{flushleft}
\end{table}
For all models, the happiness class has the highest recall at 98.15\%, while the anger class has the lowest.
Even the best model, the MLP, classified less than half of its samples correctly.
Similarly, the anger class has the lowest F-score across all models, consistently followed by the disgust and fear classes, with F-scores ranging from 51.67 to 58.41\% and 59.26 to 68.42\%, respectively.
Additionally, all models' most frequent error is misclassifying anger samples as disgust, which occurred for at least 22 out of 54 anger samples.
We identify two primary reasons for this.
First, anger and disgust are not as easily distinguishable as other expressions like happiness and fear because both involve similar-looking facial movements.
Second, anger and disgust are the least represented classes in the training set, with 66 and 102 samples, respectively. 

\subsection{Comparison with Image-based Model} \label{subsec:comp}
\begin{figure}[!hb]
  \centering
  \includegraphics[width=\linewidth]{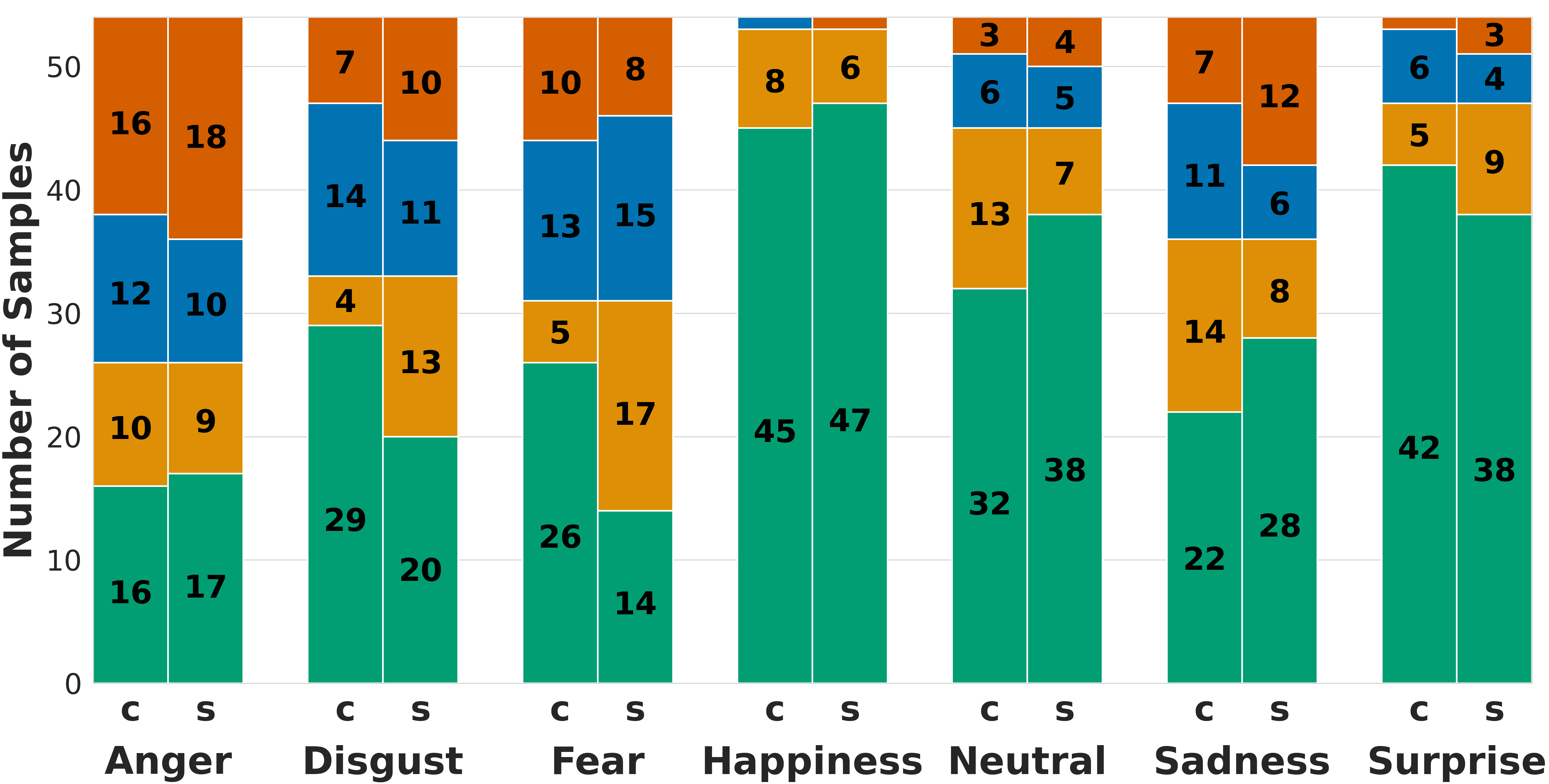}
  \caption{Agreement between our FEA-based MLP and our image-based EfficientNet-B0 \cite{ortmannEmojiherovr2024} on EmoHeVRDB's test set (c=central-view, s=side-view; colors from bottom to top: green=both correct, orange=only FEA model correct, blue=only image model correct, red=both incorrect).}
  \label{fig:pred}
\end{figure}

We experienced similar challenges with the anger and disgust classes when using EmoHeVRDB's image data \cite{ortmannEmojiherovr2024}.
However, our FEA-based approach exceeds the results of the image-based approach by up to 3.18\% in terms of accuracy.
Fig. \ref{fig:pred} presents a class-wise comparison of our MLP and EfficientNet-B0 as a stacked bar chart.
Each class's bar is divided into two sub-bars: the left represents comparisons with central-view images, while the right represents comparisons with side-view images.
We selected our MLP for this comparison because we will combine it with our EfficientNet-B0 for multimodal FER in the next
section.
Comparing the orange and blue bars, we observe that our FEA-based model significantly outperforms the image-based model in the happiness, neutral, and sadness classes.
This trend reverses only for the central-view images of the disgust and fear classes. 
The anger class remains the most challenging for both models, with 34 out of 108 samples unrecognized by either.
Overall, both models recognized the correct emotion in 414 out of 756 cases and failed in 100 cases.
In the remaining 242 cases, only the FEA-based model was correct in 128 cases, while only the image-based model was correct in 114 cases.
Given the high number of cases where one model is correct and the other is not, we conclude that there is a significant potential for fusing both models or modalities in a multimodal approach. 
If a model could perfectly determine when to use the FEA-based model's prediction and when to use the image-based model's prediction, it would achieve an accuracy of 86.77\%, correctly classifying 656 of the multimodal test set's 756 samples.

\section{Multimodal Facial Expression Recognition}
Motivated by the previous subsection's analysis of the potential for fusing modalities, we experiment with multimodal approaches in this section.
For our experiments, we relied on our EfficientNet-B0 \cite{ortmannEmojiherovr2024}, trained on EmoHeVRDB's image data, and our MLP, trained on EmoHeVRDB's FEA data.
We chose the MLP instead of the logistic regression classifier, despite its slightly worse performance, because it can be combined with the  EfficientNet-B0 into one neural network that can simultaneously process FEA and image data and learn with a combined loss function.
The MLP is sequentially composed of a Dense layer with 128 units and ReLU activation, a Dropout layer with a rate of 0.2, a Dense layer with 64 units and  ReLU activation, a Dropout layer with a rate of 0.2 and a final Dense layer with 7 units and softmax activation.

\subsection{Database Preparation}
For each of the 1,727 FEA samples we used in the previous section, EmoHeVRDB contains two simultaneously recorded images, one from a central view and the other from a 45° side view.
Thus, we combined each FEA sample with its two corresponding image samples, resulting in 3,454 multimodal samples with the same class distribution and training, validation, and test splits as before.

\subsection{Training and Results}
We experimented with intermediate and late fusion approaches and implemented our experiments with TensorFlow, version 2.15.
For all experiments, we used the Adam optimizer and the SparseCategoricalCrossentropy loss function.
We executed our experiments on an Ubuntu 22.04 machine with an Intel i9-12900K CPU, 64GB of DDR4 RAM, and an Nvidia GeForce RTX 3090 GPU.
It had CUDA 12.2 and cuDNN 8.9 installed.

\subsubsection{Late Fusion}
For late fusion, we process each model's 7-dimensional stochastic output vector without unfreezing the models' parameters.
As the first baseline, we averaged both vectors, increasing accuracy to 76.46\%. 
Then, we tried several techniques with learnable parameters.
We started with a weighted sum and concatenation followed by a Dense layer for classification.
Next, we applied bilinear fusion, followed by flattening the combination matrix and a final Dense layer.
Finally, we achieved the best late fusion result of 78.97\% accuracy with a custom cross-attention layer. 
\begin{table}[ht]
\centering
\normalsize
\caption{Performance Metrics in \% - Late Fusion }
\begin{tabular}{|p{0.16\linewidth}|P{0.15\linewidth}|P{0.15\linewidth}|P{0.3\linewidth}|}
\hline
\textbf{Emotion} & \textbf{Precision} & \textbf{Recall} & \textbf{F1} \\
\hline
Anger      & 67.01 & 60.19 & 63.41 (+ 6.27)\\
Disgust    & 60.77 & 73.15 & 66.39 (+ 9.00) \\
Fear       & 81.19 & 75.93 & 78.47 (+17.69)\\
Happiness  & 80.30 & 98.15 & 88.33 (+ 0.73)\\
Neutral    & 86.36 & 87.96 & 87.16 (+11.53)\\
Sadness    & 93.90 & 71.30 & 81.05 (+ 3.63)\\
Surprise   & 89.42 & 86.11 & 87.74 (+ 6.00)\\
\hline
\textbf{Average} & 79.85 & 78.97 & 78.94 (+ 7.84) \\
\hline
\end{tabular}
\label{table:latefusion}
\begin{flushleft}
\footnotesize
The values in parentheses after the F-scores are the differences from the results of the FEA-based MLP listed in Table \ref{table:performance_metrics} and visualized in Fig. \ref{fig:pred}.
\end{flushleft}
\end{table}
It comprised two Dense layers with seven units and softmax activation to compute weights for each model's output vector.
Each layer learns based on one output vector to weigh the other.
Afterwards, the weighted vectors are summed.
Table~\ref{table:latefusion} provides a detailed analysis of the model's performance.
Compared to our unimodal FEA-based MLP, recall values for the fear, disgust and anger classes improved by 18.52\%, 12.04\% and 12.04\%, respectively.
These findings align with our analysis in subsection III-D, where the fear, disgust and anger classes showed the highest number of samples correctly recognized only by the image-based model.
In the neutral and surprise classes, improvements were primarily achieved through increased precision, with gains of 17.13\% and 12.37\%, respectively.

\subsubsection{Intermediate Fusion}
For intermediate fusion, we relied on the EfficientNet-B0's final GlobalAveragePooling layer, which extracts 1,280 features from an input image, and on the MLP's first Dense layer, which extracts 128 features from each FEA input.
Again, we did not unfreeze the models but used them as feature extractors.
Similar to our late fusion approach, we experimented with several techniques to fuse the extracted feature vectors.
However, in contrast to late fusion, the vectors are not the same size, but the image feature vector is ten times larger than the FEA feature vector.
During our experiments, we found that scaling the feature vectors to a similar size via a subsequent Dense layer for each vector improves performance and stabilizes the learning process.
A size of 128 to 512 units worked well in most cases.
Additionally, we introduced BatchNormalization layers to ensure both feature vectors' values lie in similar ranges and increased the batch size to 128 for better mean and variance estimates and more stable training.
Then, to fuse the batch-normalized feature vectors of equal size, we experimented with concatenation, bilinear combination, and cross-attention followed by several Dense layers and intermediate Dropout layers.
The highest accuracy of 80.42\% was achieved by a model that first scaled both feature vectors to a size of 512, followed by batch normalization. 
Then, the vectors were weighted with a custom cross-attention layer, concatenated, and connected to a Dropout layer with a rate of 0.4 before the final Dense layer with seven units.
The detailed performance metrics are listed in Table \ref{table:interfusion}.
\begin{table}[ht]
\centering
\normalsize
\caption{Performance Metrics in \% - Intermediate Fusion}
\begin{tabular}{|p{0.16\linewidth}|P{0.15\linewidth}|P{0.15\linewidth}|P{0.3\linewidth}|}
\hline
\textbf{Emotion} & \textbf{Precision} & \textbf{Recall} & \textbf{F1} \\
\hline
Anger      & 75.28 & 62.04 & 68.02 (+10.88) \\
Disgust    & 66.38 & 71.30 & 68.75 (+11.36) \\
Fear       & 76.52 & 81.48 & 78.92 (+18.14) \\
Happiness  & 80.92 & 98.15 & 88.70 (+ 1.10) \\
Neutral    & 84.75 & 92.59 & 88.50 (+12.87) \\
Sadness    & 91.11 & 75.93 & 82.83 (+ 5.41) \\
Surprise   & 90.72 & 81.48 & 85.85 (+ 4.11) \\
\hline
\textbf{Average} & 80.81 & 80.42 & 80.22 (+ 9.12) \\
\hline
\end{tabular}
\label{table:interfusion}
\begin{flushleft}
\footnotesize
The values in parentheses after the F-scores are the differences from the results of the FEA-based MLP listed in Table \ref{table:performance_metrics} and visualized in Fig. \ref{fig:pred}.
\end{flushleft}
\end{table}
Compared to the late fusion model, the intermediate fusion model primarily increases precision for the anger and disgust classes and recall for the fear, neutral and sadness classes.
However, almost all changes are minor improvements or declines of under 5\%.
Nevertheless, our results demonstrate that intermediate fusion can further enhance model performance by combining data on a feature level, enabling models to learn more complex interdependencies.

\section{Conclusion and Future Work}
In this study, we explored the potential of utilizing FEAs captured via the Meta Quest Pro VR headset for static FER.
Leveraging EmoHeVRDB's FEA data, we experimented with several unimodal approaches.
Our best model, a logistic regression classifier, achieved 73.02\% accuracy, exceeding our previous results on EmoHeVRDB's image data by 3.18\%.
Furthermore, we demonstrated the potential of multimodal FER in VR settings by combining EmoHeVRDB's image and FEA data in late and intermediate fusion approaches.
An intermediate fusion model achieved the highest accuracy of 80.42\%, surpassing unimodal image-based and FEA-based models by 10.58\% and 7.40\%, respectively.
This study is the first to establish benchmarks for unimodal and multimodal FER with EmoHeVRDB's unique FEA data, marking a significant contribution to FER in VR settings.
Our code detailing the experiments conducted is available on GitHub:\linebreak
\textit{\url{https://github.com/thorbenortmann/emohevrdb-sfer}}.
For future work, we plan to explore EmoHeVRDB's sequential data for dynamic FER, aiming to contribute further to reliable FER and emotion recognition in VR settings.

\bibliographystyle{IEEEtran}

\begin{thebibliography}{10}
\providecommand{\url}[1]{#1}
\csname url@samestyle\endcsname
\providecommand{\newblock}{\relax}
\providecommand{\bibinfo}[2]{#2}
\providecommand{\BIBentrySTDinterwordspacing}{\spaceskip=0pt\relax}
\providecommand{\BIBentryALTinterwordstretchfactor}{4}
\providecommand{\BIBentryALTinterwordspacing}{\spaceskip=\fontdimen2\font plus
\BIBentryALTinterwordstretchfactor\fontdimen3\font minus \fontdimen4\font\relax}
\providecommand{\BIBforeignlanguage}[2]{{%
\expandafter\ifx\csname l@#1\endcsname\relax
\typeout{** WARNING: IEEEtran.bst: No hyphenation pattern has been}%
\typeout{** loaded for the language `#1'. Using the pattern for}%
\typeout{** the default language instead.}%
\else
\language=\csname l@#1\endcsname
\fi
#2}}
\providecommand{\BIBdecl}{\relax}
\BIBdecl

\bibitem{somarathnaVirtualRealityEmotion2022}
R.~Somarathna, T.~Bednarz, and G.~Mohammadi, ``Virtual {{Reality}} for {{Emotion Elicitation}} \textendash{} {{A Review}},'' \emph{IEEE Trans. Affective Comput.}, pp. 1--21, 2022.

\bibitem{halbigSystematicReviewPhysiological2021}
A.~Halbig and M.~E. Latoschik, ``A {{Systematic Review}} of {{Physiological Measurements}}, {{Factors}}, {{Methods}}, and {{Applications}} in {{Virtual Reality}},'' \emph{Front. virtual real.}, vol.~2, 2021.

\bibitem{marin-moralesEmotionRecognitionImmersive2020}
J.~{Mar{\'i}n-Morales}, C.~Llinares, J.~Guixeres, and M.~Alca{\~n}iz, ``Emotion {{Recognition}} in {{Immersive Virtual Reality}}: {{From Statistics}} to {{Affective Computing}},'' \emph{Sensors}, vol.~20, no.~18, p. 5163, Jan. 2020.

\bibitem{ekman2006darwin}
P.~Ekman, \emph{Darwin and facial expression: A century of research in review}.\hskip 1em plus 0.5em minus 0.4em\relax Ishk, 2006.

\bibitem{ortmannFacialEmotionRecognition2023}
T.~Ortmann, Q.~Wang, and L.~Putzar, ``Facial {{Emotion Recognition}} in {{Immersive Virtual Reality}}: {{A Systematic Literature Review}},'' in \emph{Proceedings of the 16th {{International Conference}} on {{PErvasive Technologies Related}} to {{Assistive Environments}}}, ser. {{PETRA}} '23.\hskip 1em plus 0.5em minus 0.4em\relax {New York, NY, USA}: {Association for Computing Machinery}, Aug. 2023, pp. 77--82.

\bibitem{ortmannEmojiherovr2024}
------, ``{{EmojiHeroVR}}: {{A Study on Facial Expression Recognition under Partial Occlusion from Head-Mounted Displays}},'' in \emph{2024 12th {{International Conference}} on {{Affective Computing}} and {{Intelligent Interaction}} ({{ACII}})}, 2024, preprint available on arXiv: \textit{\href{https://doi.org/10.48550/arXiv.2410.03331}{https://doi.org/10.48550/arXiv.2410.03331}}.

\bibitem{philippSocialityFacialExpressions2012}
M.~C. Philipp, K.~R. Storrs, and E.~J. Vanman, ``Sociality of facial expressions in immersive virtual environments: {{A}} facial {{EMG}} study,'' \emph{Biological Psychology}, vol.~91, no.~1, pp. 17--21, Sep. 2012.

\bibitem{pallaviciniVirtualRealityAlways2013}
F.~Pallavicini, P.~Cipresso, S.~Raspelli, A.~Grassi, S.~Serino, C.~Vigna, S.~Triberti, M.~Villamira, A.~Gaggioli, and G.~Riva, ``Is virtual reality always an effective stressors for exposure treatments? some insights from a controlled trial,'' \emph{BMC Psychiatry}, vol.~13, no.~1, p.~52, Feb. 2013.

\bibitem{bianFrameworkPhysiologicalIndicators2016}
Y.~Bian, C.~Yang, F.~Gao, H.~Li, S.~Zhou, H.~Li, X.~Sun, and X.~Meng, ``A framework for physiological indicators of flow in {{VR}} games: Construction and preliminary evaluation,'' \emph{Personal and Ubiquitous Computing}, vol.~20, no.~5, pp. 821--832, Oct. 2016.

\bibitem{chiricoEffectivenessImmersiveVideos2017}
A.~Chirico, P.~Cipresso, D.~B. Yaden, F.~Biassoni, G.~Riva, and A.~Gaggioli, ``Effectiveness of {{Immersive Videos}} in {{Inducing Awe}}: {{An Experimental Study}},'' \emph{Scientific Reports}, vol.~7, no.~1, p. 1218, Apr. 2017.

\bibitem{granatoEmpiricalStudyPlayers2020a}
M.~Granato, D.~Gadia, D.~Maggiorini, and L.~A. Ripamonti, ``An empirical study of players' emotions in {{VR}} racing games based on a dataset of physiological data,'' \emph{Multimedia Tools and Applications}, vol.~79, no.~45, pp. 33\,657--33\,686, Dec. 2020.

\bibitem{yongEmotionRecognitionGamers2019}
H.~Yong, J.~Lee, and J.~Choi, ``Emotion {{Recognition}} in {{Gamers Wearing Head-mounted Display}},'' in \emph{2019 {{IEEE Conference}} on {{Virtual Reality}} and {{3D User Interfaces}} ({{VR}})}, Mar. 2019, pp. 1251--1252.

\bibitem{georgescuRecognizingFacialExpressions2019a}
M.-I. Georgescu and R.~T. Ionescu, ``Recognizing {{Facial Expressions}} of {{Occluded Faces Using Convolutional Neural Networks}},'' in \emph{Neural {{Information Processing}}}, ser. Communications in {{Computer}} and {{Information Science}}, T.~Gedeon, K.~W. Wong, and M.~Lee, Eds.\hskip 1em plus 0.5em minus 0.4em\relax {Cham}: {Springer International Publishing}, 2019, pp. 645--653.

\bibitem{houshmandFacialExpressionRecognition2020c}
B.~Houshmand and N.~Mefraz~Khan, ``Facial {{Expression Recognition Under Partial Occlusion}} from {{Virtual Reality Headsets}} based on {{Transfer Learning}},'' in \emph{2020 {{IEEE Sixth International Conference}} on {{Multimedia Big Data}} ({{BigMM}})}, Sep. 2020, pp. 70--75.

\bibitem{georgescuTeacherStudentTraining2021}
M.-I. Georgescu, G.-E. Du{\c t}{\v a}, and R.~T. Ionescu, ``Teacher\textendash student training and triplet loss to reduce the effect of drastic face occlusion,'' \emph{Machine Vision and Applications}, vol.~33, no.~1, p.~12, Dec. 2021.

\bibitem{georgescuTeacherStudentTrainingTriplet2021a}
M.-I. Georgescu and R.~T. Ionescu, ``Teacher-{{Student Training}} and {{Triplet Loss}} for {{Facial Expression Recognition}} under {{Occlusion}},'' in \emph{2020 25th {{International Conference}} on {{Pattern Recognition}} ({{ICPR}})}, Jan. 2021, pp. 2288--2295.

\bibitem{gotsmanValenceArousalEstimation2021}
T.~Gotsman, N.~Polydorou, and A.~Edalat, ``Valence/{{Arousal Estimation}} of {{Occluded Faces}} from {{VR Headsets}},'' in \emph{2021 {{IEEE Third International Conference}} on {{Cognitive Machine Intelligence}} ({{CogMI}})}, Dec. 2021, pp. 96--105.

\bibitem{mollahosseiniAffectNetDatabaseFacial2019a}
A.~Mollahosseini, B.~Hasani, and M.~H. Mahoor, ``{{AffectNet}}: {{A Database}} for {{Facial Expression}}, {{Valence}}, and {{Arousal Computing}} in the {{Wild}},'' \emph{IEEE Trans. Affective Comput.}, vol.~10, no.~1, pp. 18--31, Jan. 2019.

\bibitem{barsoumFERPLUSTrainingDeep2016}
E.~Barsoum, C.~Zhang, C.~C. Ferrer, and Z.~Zhang, ``{{FERPLUS}} - {{Training}} deep networks for facial expression recognition with crowd-sourced label distribution,'' in \emph{Proceedings of the 18th {{ACM International Conference}} on {{Multimodal Interaction}}}, ser. {{ICMI}} '16.\hskip 1em plus 0.5em minus 0.4em\relax {New York, NY, USA}: {Association for Computing Machinery}, Oct. 2016, pp. 279--283.

\bibitem{meta_face_tracking}
{{Meta}}, ``Face tracking for {{Movement SDK}} for {{Unity}},'' Online: \textit{\href{https://developer.oculus.com/documentation/unity/move-face-tracking}{https://developer.oculus.com/documentation/unity/move-face-tracking}}, 2024, accessed: 2024-10-29.

\bibitem{meta_core_sdk}
------, ``{{Meta XR Core SDK (UPM)}},'' Online: \textit{\href{https://developers.meta.com/horizon/downloads/package/meta-xr-core-sdk/59.0}{https://developers.meta.com/horizon/downloads/package/meta-xr-core-sdk/59.0}}, 2024, accessed: 2024-10-29.

\bibitem{1370848662448368390}
P.~Ekman, ``Facial action coding system: A technique for the measurement of facial movement,'' 1978.

\end{thebibliography}

\end{document}